\documentclass[letterpaper, 10 pt, conference]{ieeeconf}  

\usepackage{amsmath}
\usepackage{amssymb}
\usepackage{mathtools}

\usepackage{tabularx}
\usepackage{tikzducks}
\usepackage{graphicx}
\usepackage{bm}
\usepackage{color}
\usepackage{float}
\usepackage{units}
\usepackage{url}
\usepackage{hyperref}
\usepackage{balance}
\usepackage{amsmath}
\usepackage{makecell}
\usepackage{cite}
\usepackage{contour}
\usepackage{xcolor}
\usepackage{amsthm}
\usepackage{algorithm, algorithmicx, algpseudocode}

\makeatletter 
\def\endfigure{\end@float} 
\def\endtable{\end@float}
\makeatother 

\usepackage{subcaption}
\usepackage[]{graphicx}
\usepackage{caption}
\IEEEoverridecommandlockouts                              
\title{\LARGE 
\textbf{On Surprising Effects of Risk-Aware Domain Randomization for Contact-Rich Sampling-based Predictive Control}
}

\author{Sergio A. Esteban$^{1}$, Junheng Li$^{1}$, Vince Kurtz$^{2}$, and Aaron D. Ames$^{1}$
\thanks{$^{1}$The authors are with the Department of Mechanical and Civil Engineering, California Institute of Technology, Pasadena, CA, USA.
    {\tt\small $\{$sesteban, junhengl, ames$\}$@caltech.edu}}%
\thanks{$^{2}$The author is with the School of Computing, DePaul University, Chicago, IL, USA. {\tt\small vkurtz1@depaul.edu}}%
\thanks{*This work was supported by Technology Innovation Institute (TII).}%
}

\usepackage{amsmath}
\usepackage{amssymb}
\usepackage{amsthm}
\usepackage{mathtools}
\usepackage{derivative}
\usepackage{xcolor}
\usepackage{bm}
\usepackage[font=small, labelfont=bf]{caption}

\usepackage{url}

\usepackage{graphicx}
\usepackage[export]{adjustbox} 
\graphicspath{{figures/}}

\usepackage{balance}

\usepackage{hyperref} 
\hypersetup{
    colorlinks=true,
    linkcolor=black,
    urlcolor=cyan,
}

\theoremstyle{definition}

\newtheorem{remark}{Remark}

\newcommand{\mc}[1]{\mathcal{#1}}

\renewcommand{\bf}{\mathbf{f}} 

\newcommand{\bp}{\mathbf{p}}

\newcommand{\bv}{\mathbf{v}}

\usepackage{blindtext}

\begin{document}

\maketitle
\thispagestyle{empty}
\pagestyle{empty}

\begin{abstract}
Domain randomization (DR) is widely used in policy learning to improve robustness to modeling error, but remains underexplored in contact-rich sampling-based predictive control (SPC), where rollout quality is highly sensitive to uncertainty. In this work, we take the first step by studying risk-aware DR in predictive sampling on a simple yet representative Push-T task, comparing average, optimistic, and pessimistic rollout aggregations under randomized model instances.
Our initial results suggest that DR affects not only robustness to model error, but also the effective cost landscape seen by the sampling-based optimizer, by reshaping the basin of attraction around contact-producing actions.
This opens up potential for exploring better grounded risk-aware contact-rich SPC under model uncertainty. Video: \url{https://youtu.be/f1F0ALXxhSM}
\end{abstract}

\section{Introduction}
Domain Randomization (DR) has become a standard and powerful tool in reinforcement learning (RL) for robotics, particularly in contact-rich settings where accurate modeling of friction, inertial parameters, and compliance remains difficult \cite{tobin2017domain}. 
In contrast, the use of DR in trajectory optimization and model predictive control (MPC) is \textit{far less} explored, especially for contact-rich problems. 
This gap is notable because the same modeling challenges that motivate DR in RL also arise in contact-rich trajectory optimization and MPC, where performance can be highly sensitive to physical parameters and contact outcomes.

Recent advances make this problem timely to study: massively parallel GPU simulators enable efficient evaluation of thousands of rollouts under different model realizations \cite{todorov2012mujoco, mittal2025isaac}, while sampling-based methods,
such as Predictive Sampling \cite{howell2022predictive}, Model Predictive Path Integral (MPPI) \cite{williams2016mppi}, and Cross Entropy Method (CEM)-style approaches \cite{rubinstein1999cross}, have become increasingly attractive for nonlinear contact-rich control because they avoid some of the analytic difficulties of the gradient-based optimization through non-smooth contact and complementarity conditions \cite{gu2025humanoid}. 
Together, these tools enable a simple paradigm: evaluate each candidate control sequence across a randomized ensemble of domains and rank it according to a chosen notion of risk.

In this work, we present a preliminary study of \textbf{risk-aware domain randomization for contact-rich sampling-based predictive control (SPC)}. In particular, our preliminary demonstration focuses on the Push-T task. 
For each candidate input sequence, we evaluate rollout cost across multiple randomized model instances and aggregate the results using three risk operators:
\begin{enumerate}
    \item \textbf{Average}: \textit{mean performance} across domains, 
    \item \textbf{Pessimistic}: \textit{worst-case} rollout cost (Risk-averse),
    \item \textbf{Optimistic}: \textit{best-case} rollout cost (Risk-seeking).
\end{enumerate}

\begin{figure}
    \centering
    \includegraphics[width=0.9\linewidth]{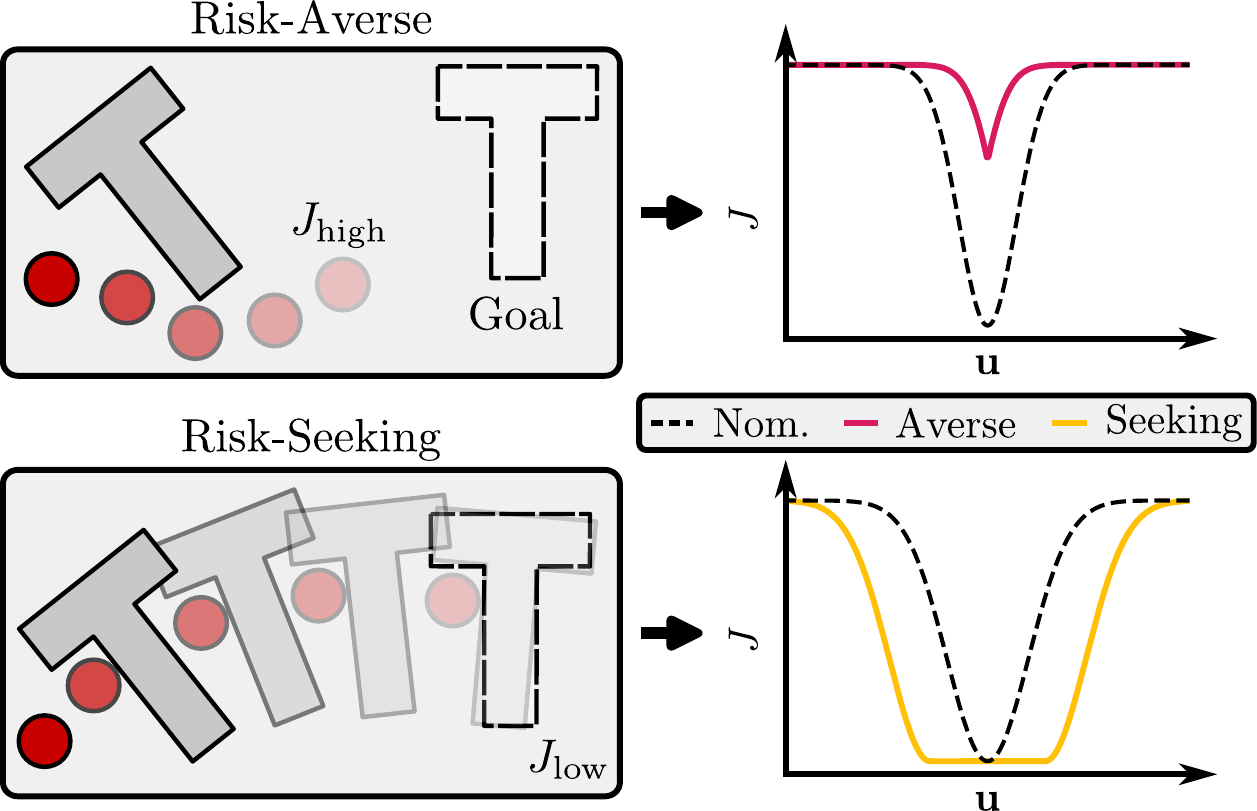}
    \caption{In contact-rich control settings, successful contact may correspond to a narrow low-cost region. We hypothesize that while risk-averse aggregation can suppress this region and shrink its basin of attraction, risk-seeking aggregation can enlarge the basin around promising contact-producing actions.}
    \label{fig:pusht}
    \vspace{-0.5cm}
\end{figure}

The main contribution of this work is a first step toward a systematic study of risk-aware DR in contact-rich sampling-based predictive control through a unified comparison of proposed risk operators. Our motivation is that the intuitive ranking of these strategies does not always hold: although pessimistic aggregation may seem most robust under model mismatch, 
our initial results suggest that optimistic aggregation can sometimes perform better by promoting high-variance but contact-producing candidates. We hypothesize that this occurs because DR affects not only robustness to modeling error, but also the effective search landscape itself. In contact-rich tasks, pessimistic aggregation can suppress narrow low-cost basins, whereas optimistic aggregation can make them easier for a sampling-based optimizer to discover. 

\section{Related Work}

Domain randomization is most established in robot learning for sim-to-real transfer, where visual or dynamics variation during training improves robustness to the reality gap \cite{tobin2017domain,andrychowicz2020learning}. Subsequent work has made DR more adaptive \cite{mehta2020active,chebotar2019closing}, but the current literature remains focused on policy learning in general \cite{muratore2022robot}.

In contrast, this work is closely related to the growing body of sampling-based predictive control methods for nonlinear planning in robotics, including MPPI \cite{williams2015model}, Predictive Sampling \cite{howell2022predictive}, and recent GPU-parallel frameworks such as Hydrax \cite{kurtz2024hydrax}. The closest prior works are robust and risk-aware MPPI variants \cite{balci2022constrained,gandhi2021robust,yin2023risk,yin2023shield,vahs2026parameter}, which mainly study robustness and safety in domains such as rally car racing or obstacle avoidance. In contrast, our focus is on contact-rich predictive control, and specifically on how risk-aware aggregation across randomized domains reshapes the search landscape of a sampling-based optimizer.

More broadly, this paper relates to robust trajectory optimization and motion planning under uncertainty, including Monte Carlo\cite{janson2017monte}, risk-bounded \cite{jasour2019risk}, scenario-based \cite{de2025scenario}, and risk-averse planning methods \cite{lew2023risk}. It is also connected to planning through contact, where recent work has used smoothing or contact-implicit formulations to address the nonsmooth, highly nonconvex structure induced by impacts and friction \cite{suh2022bundled,pang2023global,Yang-RSS-24}. Our perspective is complementary to both lines of work: rather than estimating collision risk or smoothing contact gradients, we study how risk-aware DR reshapes the cost landscape in contact-rich SPC (Fig. \ref{fig:pusht}).

\section{Sampling-based Predictive Control}

We consider a finite-horizon optimal control problem at planning time $t$, with current state estimate $\mathbf x_0=\hat{\mathbf x}(t)$. Let
\begin{align}
\mathbf U := \{\mathbf u_0, \mathbf u_1,\dots,\mathbf u_{N-1}\}
\end{align}
denote a control tape over a horizon of length $N$. The nominal planning problem can be written compactly as
\begin{align}
\min_{\mathbf U} \quad J(\mathbf U; \mathbf x_0),
\label{eq:spc_ocp}
\end{align}
where $J(\mathbf U; \mathbf x_0)$ denotes the rollout cost over the horizon under the dynamics and task objective. For contact-rich systems, this problem is often highly nonconvex and nonsmooth, making gradient-based optimization difficult. SPC instead evaluates candidate control tapes via forward rollouts and updates the nominal plan according to their costs~\cite{kurtz2025generative}.

At each MPC step, we draw $K$ samples $\mathbf{U}^{(k)}$ from some proposal distribution, typically a Gaussian centered around the previous nominal control tape. We then simulate a rollout for each sample, yielding the associated costs
\begin{align}
    J^{(k)} = J\!\left(\mathbf U^{(k)};\mathbf x_0\right), \qquad k = 1 \dots K
\label{eq:spc_nominal_cost}
\end{align}
These costs are used to update the nominal control tape $\mathbf{U}$. In general, SPC updates can be written as
\begin{align}
\mathbf U \gets \mathbf U +
\frac{\sum_{k=1}^{K} g\!\left(J^{(k)}\right)\left(\mathbf U^{(k)}-\mathbf U\right)}{\sum_{k=1}^{K} g\!\left(J^{(k)}\right)},
\label{eq:spc_weighted_update}
\end{align}
where $g:\mathbb{R}\to\mathbb{R}_{+}$ is a nonnegative weighting function. Different choices of $g(\cdot)$ recover different SPC algorithms \cite{kurtz2025generative}. In this work, we focus on predictive sampling \cite{howell2022predictive}, the simplest SPC algorithm; predictive sampling merely sets $\mathbf U$ as the lowest-cost sample.

Having updated the control tape $\mathbf U$, we apply the first action $\mathbf u_0$ and proceed in receding-horizon fashion.

\section{Risk-Aware Domain Randomization}\label{sec:risk_aware_dr}

In practice, the model we use to simulate rollouts is never perfect. Instead, we aim to improve robustness via DR. In particular, we assume that some model parameters $\boldsymbol{\theta}$ (friction coefficients, body masses, etc.) are drawn from some distribution $\mathcal{D}$. In this setting, the cost 
\begin{equation}
    J(\mathbf U; \mathbf x_0, \boldsymbol{\theta})
\end{equation}
is determined by the value of these parameters as well as the initial condition $\mathbf x_0$.

\begin{algorithm}[t]
\caption{Risk-Aware Sampling Predictive Control}
\label{alg:robust_sampling_mpc}
\begin{algorithmic}[1]
\State Initialize $\mathbf U = \{\mathbf{u}_0, \mathbf{u}_1, \dots, \mathbf{u}_N\}$
\State $\{\boldsymbol{\theta}^{(r)}\}_{r=1}^R \sim \mathcal{D}$ \Comment{sample model randomizations}
\While{planning}
    \State $\mathbf{x}_0 \gets \hat{\mathbf{x}}(t)$ \Comment update state
    \For{$k=1$ to $K$}
        \State $\mathbf{U}^{(k)} \sim \mathcal{N}(\mathbf{U}, \sigma^2$) \Comment sample control tapes
        \For{$r=1$ to $R$} 
            \State $J^{(k, r)} \gets J\!\left(\mathbf{U}^{(k)}; \mathbf{x}_0, \boldsymbol{\theta}^{(r)}\right)$ \Comment{rollouts}
        \EndFor
        \State $\bar{J}^{(k)} \gets \operatorname{Risk}\!\left(J^{(k, 1)}, J^{(k, 2)}, \dots, J^{(k, R)} \right)$
    \EndFor
    \State $\mathbf U \gets \mathbf U + \frac{\sum_{k=1}^{K} g(\bar{J}^{(k)})(\mathbf U^{(k)}-\mathbf U)}{\sum_{k=1}^{K} g(\bar{J}^{(k)})}$ \Comment{SPC step \eqref{eq:spc_weighted_update}}
    \State Apply $\mathbf u_0$
\EndWhile
\end{algorithmic}
\end{algorithm}
\vspace{-0.2cm}

\subsection{Sampled Control Inputs and Domain Randomization}

To apply DR to SPC, we randomly sample $R$ sets of parameters
\begin{equation}
    \boldsymbol{\theta}^{(r)} \sim \mathcal{D} \quad r = 1 \dots R,
\end{equation}
resulting in $R$ ``domains'', each with different dynamics. 

We roll out each control tape $\mathbf U^{(k)}$ in each domain, performing $K \times R$ total rollouts to produce costs
\begin{equation}
    J^{(k, r)} = J(\mathbf{U}^{(k)}; \mathbf x_0, \boldsymbol{\theta}^{(r)})
\end{equation}
indexed by both sample $k$ and randomized domain $r$. All $R \times K$ rollouts can be performed in parallel.

\subsection{Risk Strategies}

To apply SPC, we must aggregate the rollout costs across domains. We do so via a \textit{risk operator}
\begin{equation}
    \bar{J}^{(k)} \coloneq \operatorname{Risk}\!\left( J^{(k, 1)}, J^{(k, 2)}, \dots, J^{(k, R)}\right).
\end{equation}
This provides a single scalar score for each sample, which we can then feed into \eqref{eq:spc_weighted_update} to perform SPC.

While there are many possibilities for the \(\operatorname{Risk}(\cdot)\) operator, we focus here on three simple strategies: \textit{average}, \textit{pessimistic}, and \textit{optimistic}.

\subsubsection{Average}
Strategy that uses the average cost over the randomized domains:
\begin{equation}
    \bar{J}_{\mathrm{avg}}^{(k)}
    \approx \frac{1}{R}\sum_{r=1}^{R} J{(k, r)}.
\end{equation}
This is most similar to the DR used in RL, which optimizes performance in expectation over randomized domains.

\subsubsection{Pessimistic}
Strategy that assumes the worst, and uses the highest cost across the randomized models:
\begin{equation}
    \bar{J}_{\mathrm{pes}}^{(k)} \coloneq \max_r J^{(k,r)}.
\end{equation}
This is a risk-averse strategy: costs are only low if they are low across all domains.

\subsubsection{Optimistic}
Strategy that assumes the best, and uses the lowest cost across the randomized models:
\begin{equation}
    \bar{J}_{\mathrm{opt}}^{(k)} \coloneq \min_r J^{(k, r)}.
\end{equation}
This is a risk-seeking strategy and may seem like a strictly bad idea, especially under modeling error. However, as we show below, risk-seeking DR can produce strong and even superior performance on contact-rich tasks.
\begin{remark}
Risk metrics such as VaR and CVaR \cite{yin2023risk} can interpolate between risk-seeking (optimistic), risk-neutral (average), and risk-averse behavior (pessimistic), and are a natural direction for future study.
\end{remark}
Algorithm~\ref{alg:robust_sampling_mpc} summarizes the overall procedure. Note that both loops in this algorithm are trivially parallelizable. In practice, performant SPC implementation requires spline-based dimensionality reduction and
careful treatment of time shifts between replanning steps. We omit these details for space and refer the interested reader to \cite{howell2022predictive, kurtz2024hydrax, jordana2025introduction}.

\subsection{Implementation Details}
We implement the proposed MPC framework in Hydrax \cite{kurtz2024hydrax}, using MuJoCo MJX \cite{todorov2012mujoco} as the backend for parallel rollout evaluation across sampled input tapes and domain-randomized model instances. 
\subsection{Experiment Setup}
The Push-T task consists of two bodies: a T-shaped block and a spherical pusher, both governed by second-order mechanical dynamics. The goal is for the spherical pusher to drive the T-shaped block to a desired pose through contact. 
As illustrated in Fig.~\ref{fig:pusht}, the spherical pusher has configuration 
$\mathbf{q}_p = \mathbf{p}_p \in \mathbb{R}^2$, where $\mathbf{p}_p$ denotes its planar position. 
The T-block has configuration $\mathbf{q}_b = [\mathbf{p}_b,\phi] \in SE(2)$, 
where $\mathbf{p}_b \in \mathbb{R}^2$ denotes its planar position and $\phi \in \mathbb{S}^1$ its orientation.

We compare the three risk-aware DR strategies from Sec.~\ref{sec:risk_aware_dr} using predictive sampling \cite{howell2022predictive}, with controller parameters given in Table~\ref{tab:predictive_sampling_params} and cost terms given in Table~\ref{tab:pusht_cost_terms}. To model parametric uncertainty, we define the DR terms as
\begin{equation}
    \boldsymbol{\theta} =
    \{\boldsymbol{\lambda},\; \boldsymbol{\tau},\; \mathbf{m},\; \mathbf{k}_v
    \},
\end{equation}
where $\boldsymbol{\lambda}$ and $\boldsymbol{\tau}$ are the sliding-friction and contact-time-constant vectors, and $\mathbf{m}$ and $\mathbf{k}_v$ are the body-mass and actuator-gain vectors. The sampling strategy associated with each component of $\boldsymbol{\theta}$ is summarized in Table~\ref{tab:pusht_domain_randomization}.

At each MPC step, predictive sampling returns a pusher planar velocity command $\mathbf u_0$, which is applied to the pusher through a first-order closed-loop velocity servo:
\begin{align}
    \mathbf{v}_p = \dot{\mathbf{p}}, \quad \dot{\mathbf{v}}_p=\mathbf k_v(\mathbf u_0- \mathbf{v}_p).
\end{align}
%
To evaluate robustness to model mismatch, each trial is associated with a unique seed that specifies both the collision-free initial condition (Table \ref{tab:initial_state_sampling}) and a fixed “true” model realization sampled from the randomization distribution. At planning time, the controller evaluates candidate actions over an ensemble of $R$ randomized models, whereas execution takes place on the single fixed true model realization, thereby introducing model mismatch. We consider $R \in \{0, 4, 16, 32, 64\}$ together with the three risk strategies. For $S=20$ seeded trials, we simulate for $T_{\mathrm{sim}} = 7.0$ seconds.

\begin{table}[!t]
    \centering
    \caption{Predictive sampling parameters used in the experiments.}
    \begin{tabular}{c c c c c}
        \hline
        \textbf{Samples} & \textbf{Noise} & \textbf{Horizon} & \textbf{Spline} & \textbf{Knots} \\
        \hline
        $K = 128$ & $\sigma = 0.4$ & $T = 0.5$ sec & Zero-order & $n = 6$ \\
        \hline
    \end{tabular}
    \label{tab:predictive_sampling_params}
    \vspace{-0.25cm}
\end{table}

\begin{table}[!t]
    \centering
    \caption{Cost terms used in the experiments, where $\|\cdot\|$ is the Euclidean norm and $\ominus$ denotes the orientation difference operator.}
    \begin{tabular}{c c c}
        \hline
        \textbf{Cost Term} & \textbf{Weight} & \textbf{Cost Function} \\
        \hline
        Block Position & $w_p = 2.0$ & $\|\bp_{b}^{\rm des} - \bp_{b}\|^2$ \\
        Block Orientation & $w_q = 1.0$ & $\|\phi^{\rm des} \ominus \phi\|^2$ \\
        Pusher Close-to-Block & $w_c = 0.01$ & $\|\bp_{p} - \bp_{b}\|^2$ \\
        Pusher Velocity & $w_v = 0.01$ & $\|\bv_{p}\|^2$ \\
        \hline
    \end{tabular}
    \label{tab:pusht_cost_terms}
    \vspace{-0.1cm}
\end{table}

\begin{table}[!t]
    \centering
    \caption{DR variables for the task. Contact parameters are sampled directly, while body mass and actuator velocity gain are multiplicatively scaled. $\mc{U}(\cdot, \cdot)$ denotes the uniform distribution and $\odot$ denotes the Hadamard product for element-wise multiplication.}
    \begin{tabular}{c c}
        \hline
        \textbf{Randomization Variable} & \textbf{Application} \\
        \hline
        Sliding Friction & $\boldsymbol{\lambda} \sim \mathcal{U}(0.5,\;1.5)$ \\
        Contact Time Const. & $\boldsymbol{\tau} \sim \mathcal{U}(0.01,\;0.03)$ \\
        Body Mass & $\mathbf{s}_m \odot \mathbf{m},\quad \mathbf{s}_m \sim \mathcal{U}(0.8,\;1.2)$ \\
        Actuator Velocity Gain & $\mathbf{s}_k \odot \mathbf{k}_v,\quad \mathbf{s}_k \sim \mathcal{U}(0.8,\;1.2)$ \\
        \hline
    \end{tabular}
    \label{tab:pusht_domain_randomization}
    \vspace{-0.25cm}
\end{table}

\begin{table}[!t]
    \centering
    \caption{Initial-condition sampling used in the Push-T experiments. All initial velocities are zero.}
    \begin{tabular}{c c c}
        \hline
        \textbf{T-block position} & \textbf{T-block orientation} & \textbf{Pusher position} \\
        \hline
        $\mathbf{p}_b \sim \mathcal{U}(-0.1,\;0.1)$ &
        $\phi \sim \mathcal{U}(-3.14,\;3.14)$ &
        $\mathbf{p}_p \sim \mathcal{U}(-0.1,\;0.1)$ \\
        \hline
    \end{tabular}
    \label{tab:initial_state_sampling}
    \vspace{-0.25cm}
\end{table}

\subsection{Results}
\begin{figure*}[!t]
    \centering
    \begin{subfigure}[t]{0.49\textwidth}
        \centering
        \includegraphics[width=0.87\linewidth]
        {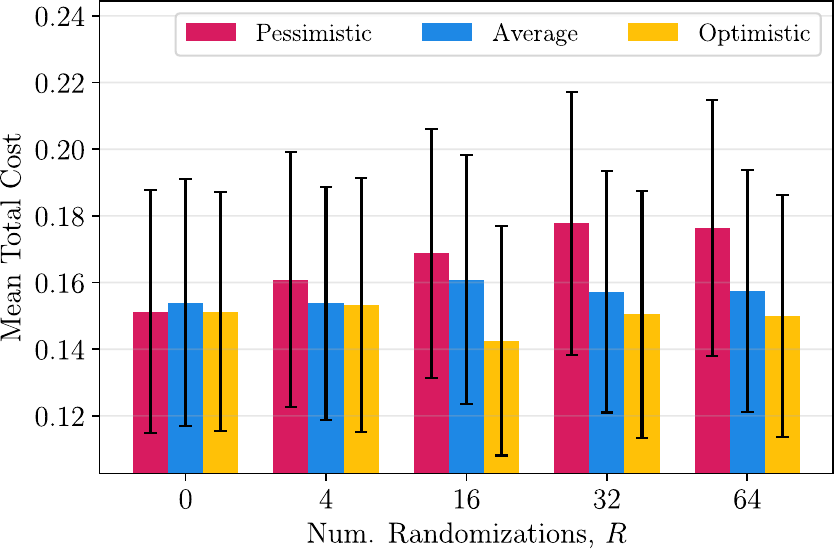}
        \caption{}
        \label{fig:cost_bar_graph}
    \end{subfigure}
    \hfill
    \begin{subfigure}[t]{0.47\textwidth}
        \centering
        \includegraphics[width=\linewidth]
        {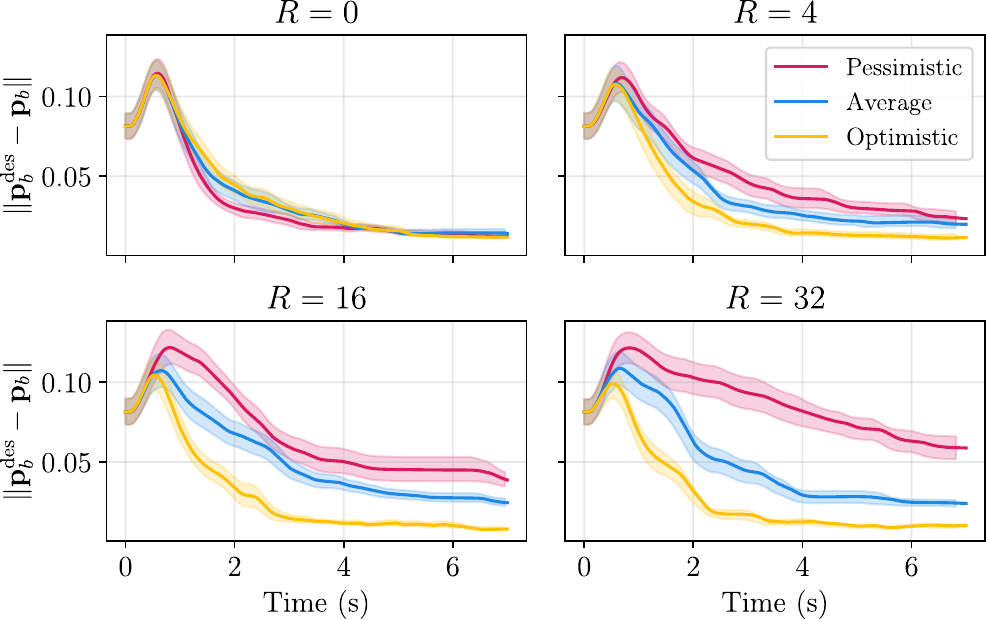}
        \caption{}
        \label{fig:cost_time_curve}
    \end{subfigure}
    \vspace{-5.0 pt}
    \caption{Comparison of risk-sensitive domain randomization strategies on the Push-T task, averaged over $S=20$ simulations with distinct randomization seeds. (a) Time-averaged total cost over $T_{\mathrm{sim}}=7.0$ second trajectories ($\pm$ SE). (b) Block position error over time ($\pm$ SE) for selected values of $R$.}
    \label{fig:main_results}
    \vspace{-0.2cm}
\end{figure*}

All reported metrics are computed on the executed closed-loop simulation rather than on the predicted rollouts, so the risk operator affects performance only indirectly through the planner’s sample ranking. Fig. \ref{fig:cost_bar_graph} shows that the \textit{pessimistic} strategy generally performs worst and degrades as the number of randomized domains increases, while the average strategy remains intermediate. 
In contrast, the \textit{optimistic} strategy consistently achieves the best or near-best mean total cost, with the strongest performance around $R=16$.
Fig. \ref{fig:cost_time_curve} shows a similar trend in block position error over time. As $R$ increases, the \textit{optimistic} strategy drives the block toward the goal more quickly, whereas the \textit{pessimistic} strategy often stalls and fails to make meaningful progress. Accordingly, in difficult Push-T cases, the \textit{optimistic} controller reaches the low-cost target region much earlier, whereas the \textit{pessimistic} controller often fails to do so. 

\section{Discussion} \label{sec:discussion}

\begin{figure}
    \centering
    \includegraphics[width=0.95\linewidth]{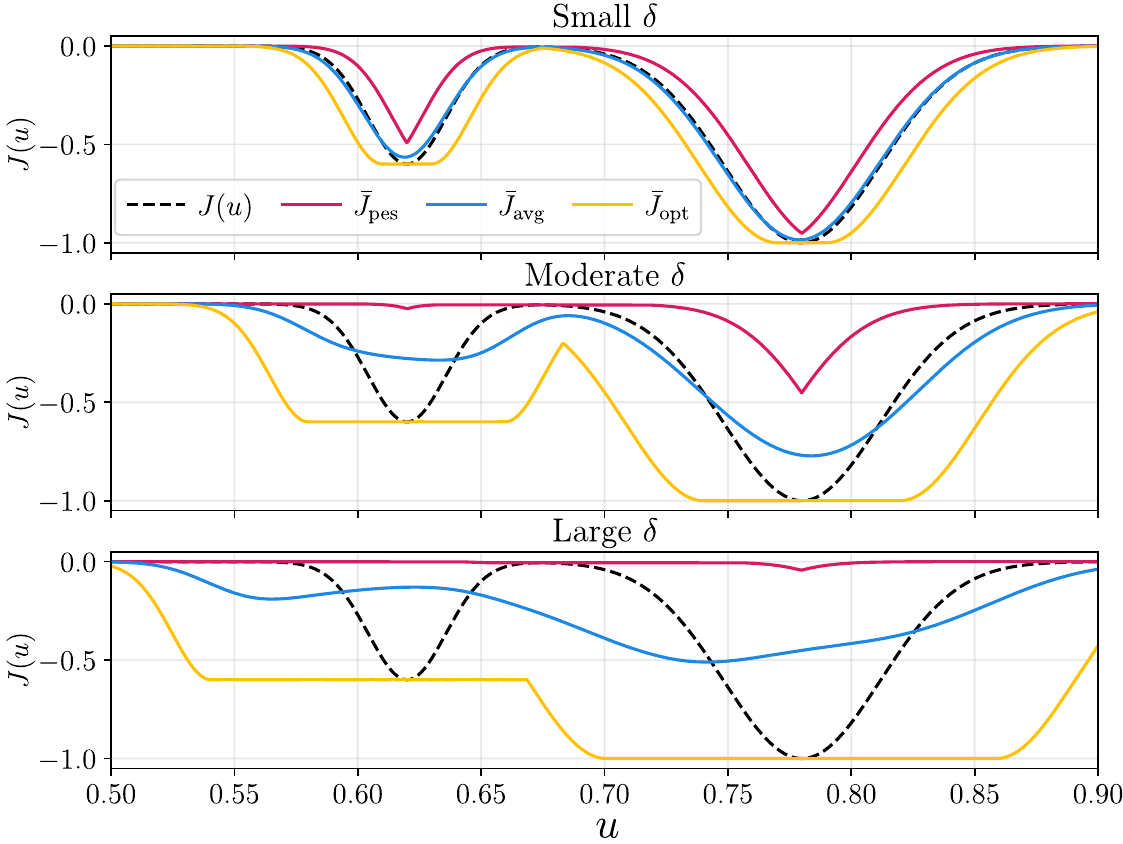}
    \caption{Effect of risk-sensitive domain randomization on a scalar cost landscape $J(u)$ with two local minima. Each row corresponds to a different perturbation magnitude $\delta$. The optimistic strategy $\bar{J}_{\mathrm{opt}}$ widens the basins of attraction around local minima, while the pessimistic strategy $\bar{J}_{\mathrm{pes}}$ narrows them. As $\delta$ increases, these basin-shaping effects become more pronounced.}
    \label{fig:cost_function_landscape}
    \vspace{-0.6cm}
\end{figure}
Intuitively, one might expect a risk-averse strategy to perform best under model error. But this expectation is not supported by data, at least in the case of the Push-T. In fact, the trend that we see in Fig.~\ref{fig:main_results} is quite the opposite: the risk-\emph{seeking} strategy instead achieves the best performance.

We hypothesize that this surprising result is due to the fact that domain randomization plays two unique and sometimes conflicting roles: (i) improving robustness to model error, and (ii) shaping the basin of attraction around local minima. The first role is well-known, while the second is less so. 

To illustrate the basin-shaping effect, consider a simple scalar cost $J(u)$. Domain randomization warps the cost landscape in complicated ways, but for ease of illustration, assume that DR merely shifts the cost landscape left and right. More precisely, for each randomized domain $r$ we have
\begin{equation}
    J^{(r)}(u) = J(u + \epsilon), \quad \epsilon \in \mathcal{U}(-\delta, \delta).
\end{equation}
An example is shown in Fig.~\ref{fig:cost_function_landscape}, along with aggregated costs $\bar{J}_{\mathrm{avg}}$, $\bar{J}_{\mathrm{pes}}$, $\bar{J}_{\mathrm{opt}}$ from each of our three risk strategies. 

The risk-seeking optimistic strategy ($\bar{J}_{\mathrm{opt}}$) expands the basin of attraction around local minima, while the risk-averse pessimistic strategy ($\bar{J}_{\mathrm{pes}}$) makes local minima smaller and narrower. Because contact-rich tasks are dominated by large (nearly) flat regions and narrow minima, clearer information about the location of a local minimum, as provided by a risk-seeking strategy, may be more important than preferring a very robust local minima. However, a risk-seeking strategy can also degrade performance by obscuring the location of a true minimum. This is quite similar to randomized smoothing \cite{suh2022bundled}, though DR perturbs the cost function itself rather than simply adding noise to its arguments.

This effect can be seen from a simple rollout-ranking example in Push-T.
Suppose candidate \(A\) has randomized costs \(\{1,8,9\}\), while candidate \(B\) has costs \(\{4,4,4\}\), where lower is better. Candidate \(B\) is consistent but mediocre, and is preferred by pessimistic aggregation as \(\max (A)>\max(B)\). Candidate \(A\), however, succeeds in one randomized domain and is preferred by optimistic aggregation as \(\min(A)<\min(B)\).
Fig.~\ref{fig:pusht} illustrates such a scenario. Since local minima are particularly narrow and difficult to hit in contact-rich tasks like the Push-T, taking this risky rollout is often worthwhile; it moves the system closer to a local minimum, where further replanning steps will be more effective.

\section{Future Work and Open Questions}

The work presented here is extremely preliminary---a single task, one SPC algorithm, and only a few risk strategies. Future experimental work will focus other contact-rich tasks, more sophisticated SPC algorithms (MPPI, CEM, CMA-ES, etc.), more advanced risk strategies (VaR, CVaR, etc.), and sim-to-real transfer.

Our preliminary results also point to a pressing need for more fundamental understanding of risk-aware domain randomization for contact-rich control. What theoretical framework(s) should we leverage? Can we quantify the tradeoff between model robustness and a friendlier cost landscape? Do such tradeoffs occur in policy learning as well as predictive control? How do system dynamics (quasi-static vs unstable, high-dimensional vs low-dimensional, etc.) influence these tradeoffs? And can we use this information to design better SPC algorithms?

\section{Conclusion}

Thanks to recent advances in hardware-accelerated parallel simulation and sampling-based optimization, we finally have the tools to investigate domain randomization in the predictive control setting. In this preliminary study, we demonstrated that risk-aware domain randomization produces some surprising and counter-intuitive effects for contact-rich tasks. While preliminary and limited in scope, these initial results demonstrate striking qualitative differences from the well-known impact of domain randomization on policy learning, pointing toward domain randomized predictive control as an important and fruitful area for further study.





\clearpage
\balance

\bibliographystyle{ieeetr}
\bibliography{References/references.bib}

\end{document}